\documentclass[conference]{IEEEtran}
\IEEEoverridecommandlockouts
\usepackage{cite}
\usepackage{amsmath,amssymb,amsfonts}
\usepackage{algorithmic}
\usepackage{graphicx}
\usepackage{textcomp}
\usepackage{xcolor}
\def\BibTeX{{\rm B\kern-.05em{\sc i\kern-.025em b}\kern-.08em
    T\kern-.1667em\lower.7ex\hbox{E}\kern-.125emX}}

\usepackage{graphicx}
\usepackage{multicol}
\usepackage{multirow}
\usepackage{booktabs}
\usepackage{wrapfig}
\usepackage{hyperref}
\usepackage[table]{xcolor}
\usepackage[ruled,vlined,linesnumbered]{algorithm2e}
\usepackage{flushend}

\begin{document}

\title{Detection of Extremely Rare Dead Cells in Fluorescence Microscopy Images}

\author{\IEEEauthorblockN{Prateek Verma$^1$, Benjamin Butter$^2$, Jian Zhang$^2$, Xintao Wu$^1$}
$^1$\textit{Department of Electrical Engineering and Computer Science}\\
$^2$\textit{Department of Biomedical Engineering}\\
University of Arkansas, Fayetteville, Arkansas, USA\\
\{prateek, bbutter, jianz, xintaowu\} @ uark.edu}

\maketitle


\begin{abstract}
Dead cell detection in fluorescence microscopy is essential for phenotypic screening and drug efficacy assessment. However, automated identification remains challenging due to signal variability, optical artifacts, and colocalization ambiguities across multiple channels. Markers/stains that define nuclear boundaries do not neatly, in practice, coincide spatially or temporally with apoptotic markers. Furthermore, marker accumulation kinetics vary across individual cells and apoptosis stages, making universal thresholds unreliable. We present a systematic comparison of several computational approaches for detecting apoptotic cells in nucleus-stained specimens: (1) heuristic methods using intensity thresholds, distributions, correlations, and domain guidance; (2) classification using both numerical features (Logistic Regression, Random Forest, XGBoost and LightGBM) and image features (ResNet-18 and EfficientNet-B0); and (3) anomaly detection, also using both numerical features (Isolation Forest, one-class SVM and PCA reconstruction) and image features (convolutional autoencoders and Deep SVDD). Using a dataset of 258 images containing 65,527 cells/nuclei across multiple scenes with 243 manually annotated ground truth dead cells, we evaluate each method on metrics such as PR-AUC, Precision, Recall, F1-score, and macro-F1. Our results indicate that classification using image features (deep learning) and numerical features (traditional) achieved the best performance, with ResNet-18 and XGBoost significantly outperforming manually designed heuristics and anomaly detection methods. Our findings demonstrate that such biomedical problems remain highly challenging to solve even with the state-of-the-art computational and learning techniques. We aim to provide an effective decision-support tool to researchers while tackling such problems either fully-automatically or in a human-in-the-loop setting as an aide to domain scientists.
\end{abstract}



\section{Introduction}

\begin{figure*}[!ht]
    \centering
    \includegraphics[width=0.95\linewidth]{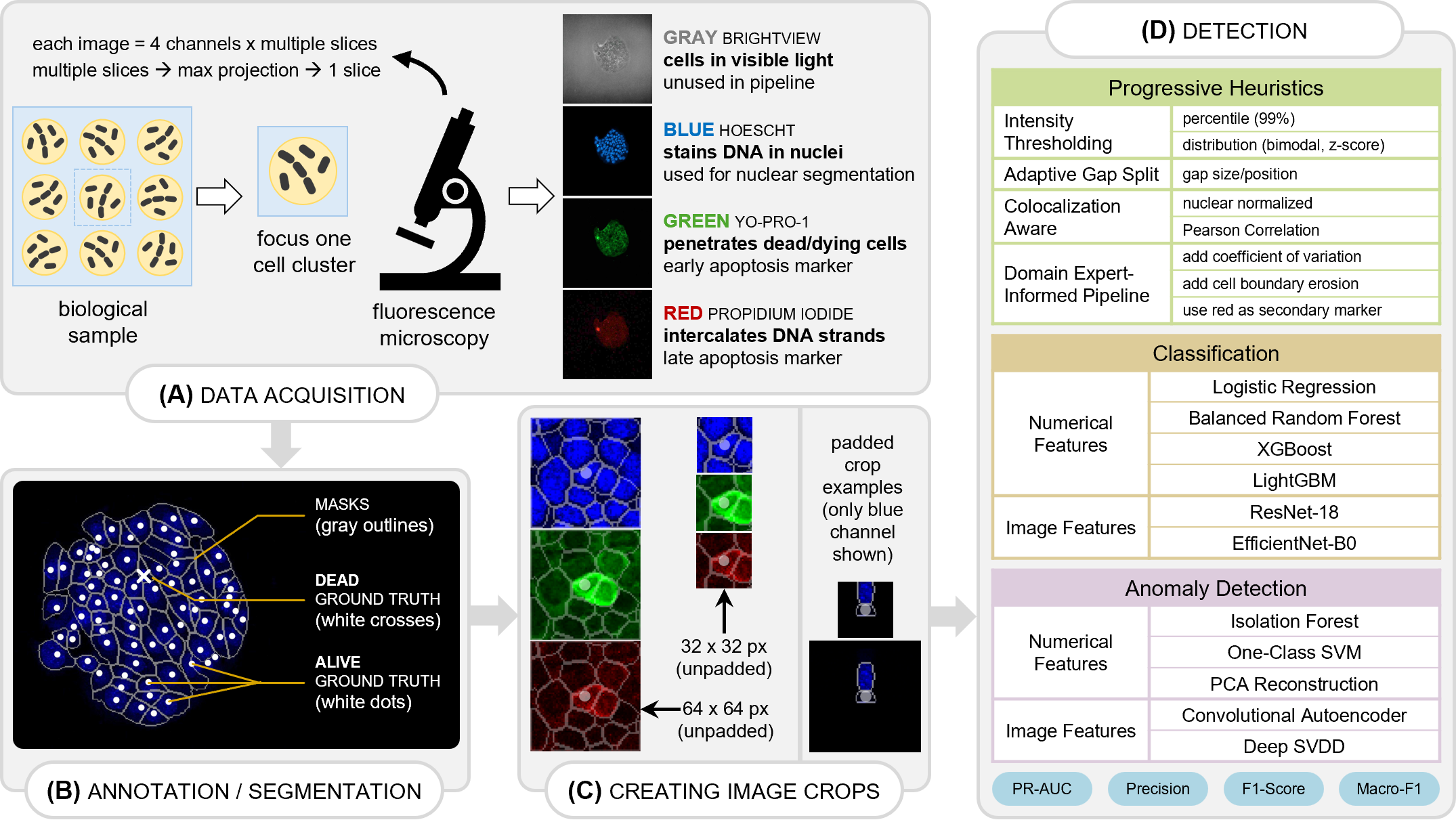}
    \caption{An illustration of the computational approaches used in the automation pipeline.}
    \label{fig:illustration}
\end{figure*}

Cell viability assessment is critical across multiple biomedical disciplines, including drug discovery, toxicology, and regenerative medicine. Manual inspection of apoptotic markers in fluorescence microscopy images is time-consuming, subjective, and impractical for high-throughput screening campaigns. A primary approach to dead cell detection leverages colocalization of nuclear and apoptotic markers. The hypothesis is that apoptotic cells accumulate specific fluorophores in the nucleus region (typically marked by a blue DNA stain like Hoechst or DAPI); detecting simultaneous enrichment in both channels provides specificity \cite{galluzzi2018molecular, wlodkowic2011apoptosis, chazotte2011labeling}.

To identify dead cells, domain scientists visually inspect the colocalization of fluorescence signals, paying close attention when a green apoptosis signal (like that from YO-PRO-1) overlaps with a nucleus identified by the blue channel (like that from Hoechst). Hoechst works by staining DNA in the nuclei of healthy cells and YO-PRO-1 by penetrating the cell membrane of dying cells and staining the double stranded nucleic acids. Because of the underlying biological mechanics, a valid target signal must match the morphology of the nuclear topology, meaning any hyper-intensity that does not follow the nuclear shape is discarded as noise or stained cell debris \cite{galluzzi2018molecular}. A secondary apoptosis signal from a red channel (like that from Propidium Iodide or simply, PI) provides a good confirmation to confidently ``double-check" these dead cells; PI works by slipping between DNA base pairs (intercalation) for already dead or damaged cells. The green channel remains the primary driver of classification because its dye stains slightly earlier in the apoptotic cascade, catching cells that have not yet accumulated the red marker \cite{wlodkowic2011apoptosis}. Translating this expert manual protocol into an automated pipeline, however, introduces severe computational challenges. The problem is characterized by an extreme class imbalance where dead cells represent a tiny fraction of the population \cite{buda2018systematic}. It is further complicated by imperfect segmentation, manual annotation limits, image-to-image variations, and staining artifacts making it highly difficult for rigid algorithms to reliably separate valid apoptotic signals from background noise.

We systematically evaluated three computational families for detecting apoptotic cells in a curated dataset with expert-annotated ground truth. First, \textbf{heuristic} based methods used simple intensity thresholds (based on percentile, bimodality or z-score), adaptive intensity gap detection algorithm to identify the natural separation between live and dead cell populations, Pearson Correlation or normalization between nuclear and apoptotic marker channels as a colocalization check, and eventually a domain-knowledge incorporated pipeline that aimed to remove blotchy/bright and edge stain artifacts and use red channel for confirmation only \cite{rousseeuw2018anomaly, balaji2024enhancing, manders1993measurement}. Second, traditional \textbf{classification} models (Logistic Regression \cite{hosmer2013applied}, Balanced Random Forest \cite{breiman2001random}, XGBoost \cite{chen2016xgboost}, and LightGBM \cite{ke2017lightgbm}) were trained on numerical features extracted from Cellpose segmentation masks and square image patches centered at the segmented nuclear centroids. Deep learning approach utilized ResNet-18 \cite{he2016deep} and EfficientNet-B0 \cite{tan2019efficientnet} architectures on the same image patches, using three-channel inputs (blue/nucleus, green/apoptotic marker, red/confirmation marker). Finally, \textbf{anomaly detection} methods, using both numerical features (Isolation Forest \cite{liu2008isolation}, one-class SVM \cite{scholkopf2001estimating}, PCA reconstruction \cite{wold1987principal}) and image features (convolutional autoencoders \cite{sakurada2014anomaly} and Deep SVDD \cite{ruff2018deep}) were trained exclusively on live cells to identify out-of-distribution samples as potential dead cells.

Our acquired dataset consisted of 112,011 segmented cells across 766 images, with 258 images having expert-annotated ground truth. Manual annotation provided 243 dead cell labels and 65,284 remaining cells classified as alive. For learning-based approaches, the data was partitioned into an 80/20 train/test split, where decision thresholds for classification models were optimized using group-aware five-fold cross-validation on the training set. Heuristic and anomaly detection approaches were applied to identical data scopes using prevalence-informed or image-specific criteria described in their respective sections. Performance across all families is uniformly reported on the evaluation partition using PR-AUC (where applicable), Precision, Recall, F1-score, and macro-F1 as primary metrics \cite{davis2006relationship, buda2018systematic, sokolova2009systematic}.


\section{Data Collection and Preparation}

\subsection{Data Acquisition and Ground Truth Annotation}

Fluorescence microscopy images were acquired from PBS-treated breast cancer cell samples cultured on glass slides using a multi-channel imaging protocol with four optical channels: gray brightview images unused in this study, blue (nucleus/Hoechst), green (apoptotic marker YO-PRO-1), and red (Propidium Iodide or PI, secondary/confirmation marker). PBS-treatment was used to induce apoptosis. Image data were stored in CZI format (Zeiss) and organized into scenes representing distinct regions of interest. Multiple image slices along the z-axis obtained during acquisition were combined into one using the maximum projections (Figure \ref{fig:illustration}a).

Manual annotation was performed by domain experts who marked dead cell locations directly on the image using ImageJ. These ground truth points were matched to Cellpose-generated masks using a 10-pixel tolerance criterion: if a point fell inside a mask, it was assigned to that cell; otherwise, the nearest cell centroid within 10 pixels (if any) was assigned. This process resulted in 243 dead cell labels mapped to segmented cells across 258 annotated images (87 could not be mapped). Alive cells were also labeled similarly by experts. However, in this study, all masks that did not map as dead cells were deemed alive and so the alive cell ground truth annotations remained unused. Table \ref{tab:data_characteristics} provides detailed counts for the data and Figure \ref{fig:illustration}b shows an example annotation.

\begin{table}[t]
\centering
\scriptsize
\caption{Data Characteristics}
\label{tab:data_characteristics}
\begin{tabular}{clr}
\toprule
\textbf{} & \textbf{} & \textbf{counts} \\
\midrule
(a) & images & 766 \\
(b) & total detected cell/nuclear masks & 112,011 \\
(c) & images with ground truth & 258 \\
(d) & masks in images with ground truth & 65,527 \\
(e) & dead cell labels & 330 \\
(f) & masks mapped to dead labels & 243 \\
(g) & masks deemed alive & 65,284 \\
\midrule
(h) & generated patches (32px, nopad) & 65,266 (243 dead) \\
(i) & generated patches (32px, pad) & 65,527 (243 dead) \\
(j) & generated patches (64px, nopad) & 64,925 (243 dead) \\
(k) & generated patches (64px, pad) & 65,527 (243 dead) \\
\bottomrule
\end{tabular}
\end{table}

In the remainder of this manuscript, we will frequently use the term `cell' when in reality it is mainly the nuclei that is being stained because the results and discussion relate to the cell as a whole. Additionally, we will frequently use the color channel names (blue, red, etc.) when referring to images instead of the fluorescent dye names for the sake of simplicity.

\subsection{Cell Segmentation}
Cells were segmented using Cellpose v4 \cite{pachitariu2025cellpose} on the blue channel, which provides robust identification of individual cells even under varying staining intensities. The resulting mask arrays assign each pixel to either background or a specific cell identity (integer label), enabling per-cell intensity measurements and spatial analysis. While a systematic sensitivity analysis of classification performance relative to segmentation error bounds falls outside the scope of this work, our model selection was informed by rigorous preliminary qualitative benchmarking that included the Segment Anything Model 2 (SAM-2). Ultimately, Cellpose v4 was selected as the core engine. As a generalist foundational model for cellular morphometry, it consistently demonstrates state-of-the-art zero-shot generalization that is typically only surpassed by specialized architectures fine-tuned extensively on domain-specific target distributions. 

\subsection{Image Patches}
\label{sec:data_prep_patches}
For deep learning classification and anomaly detection models, we extracted square image patches centered on each segmented cell centroid. We evaluated four patch variants: $32 \times 32$ pixels and $64 \times 64$ pixels in both padded (constant padding for boundary cells) and unpadded configurations. The 32 pixel patches focus mostly on the cell in consideration (as the average cell diameter was around 15) while the the 64 pixel patches bring a few neighboring cells into consideration as well. The three-channel inputs to the models were normalized by dividing by the maximum possible value for the image's bit depth (typically 16-bit), yielding values in [0, 1]. These patches were also used to yield numerical features for traditional classification and anomaly detection methods. Please see Figure \ref{fig:illustration}c for examples.

\begin{figure*}[t]
    \centering
    \includegraphics[width=1.0\linewidth]{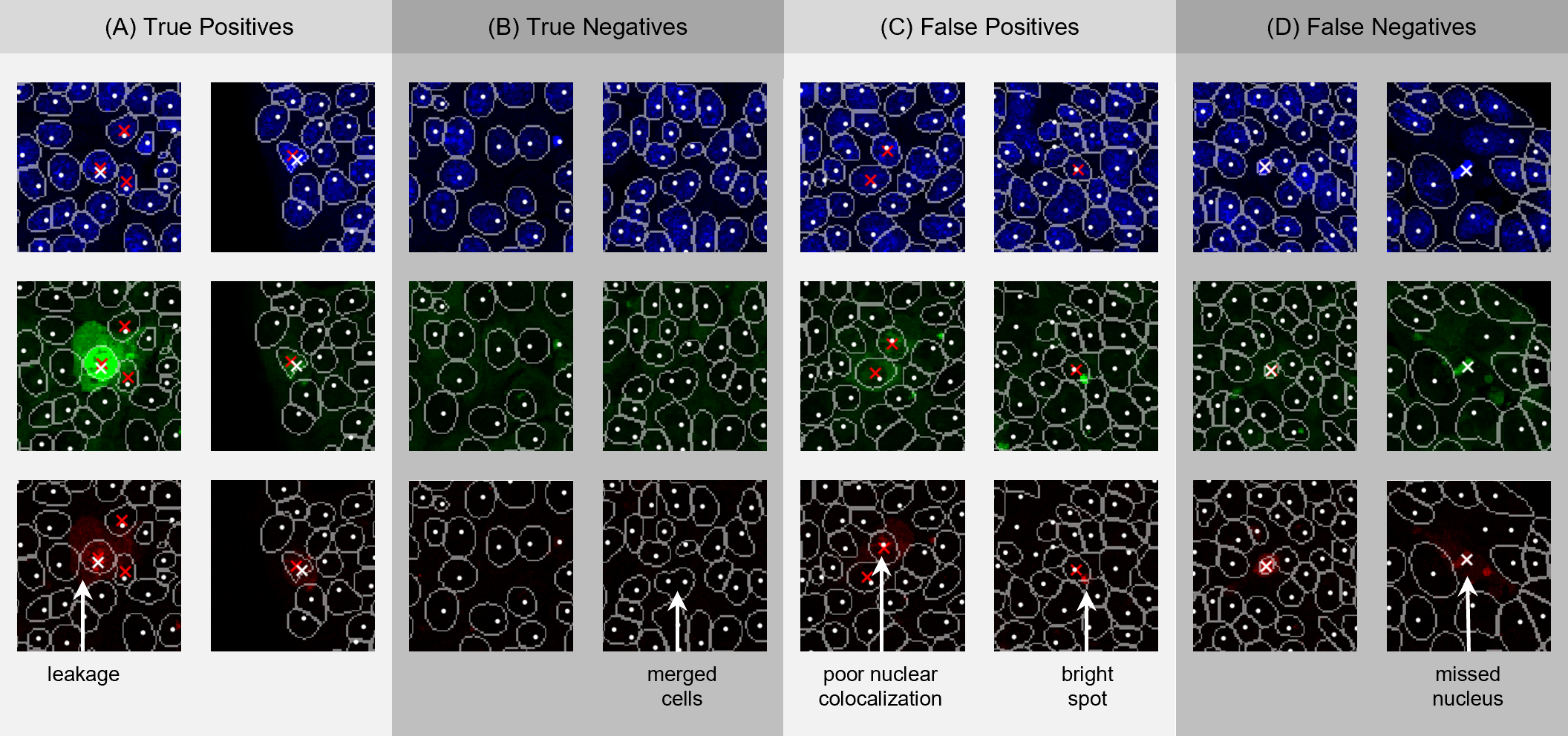}
    \caption{Two examples each of TP, TN, FP, and FN detections by the best performing \textbf{heuristic model} (percentile intensity thresholding). The cells of interest are centered in the tile. The three rows correspond to the blue, green and red channels respectively. The white outlines correspond to the Cellpose masks. White dots/crosses indicate the exact marked position of alive/dead cells as annotated by the domain expert. The red crosses, always placed at centroids of the Cellpose mask, indicate a predicted dead cell.}
    \label{fig:heuristic_results}
\end{figure*}

\begin{figure*}[t]
    \centering
    \includegraphics[width=1.0\linewidth]{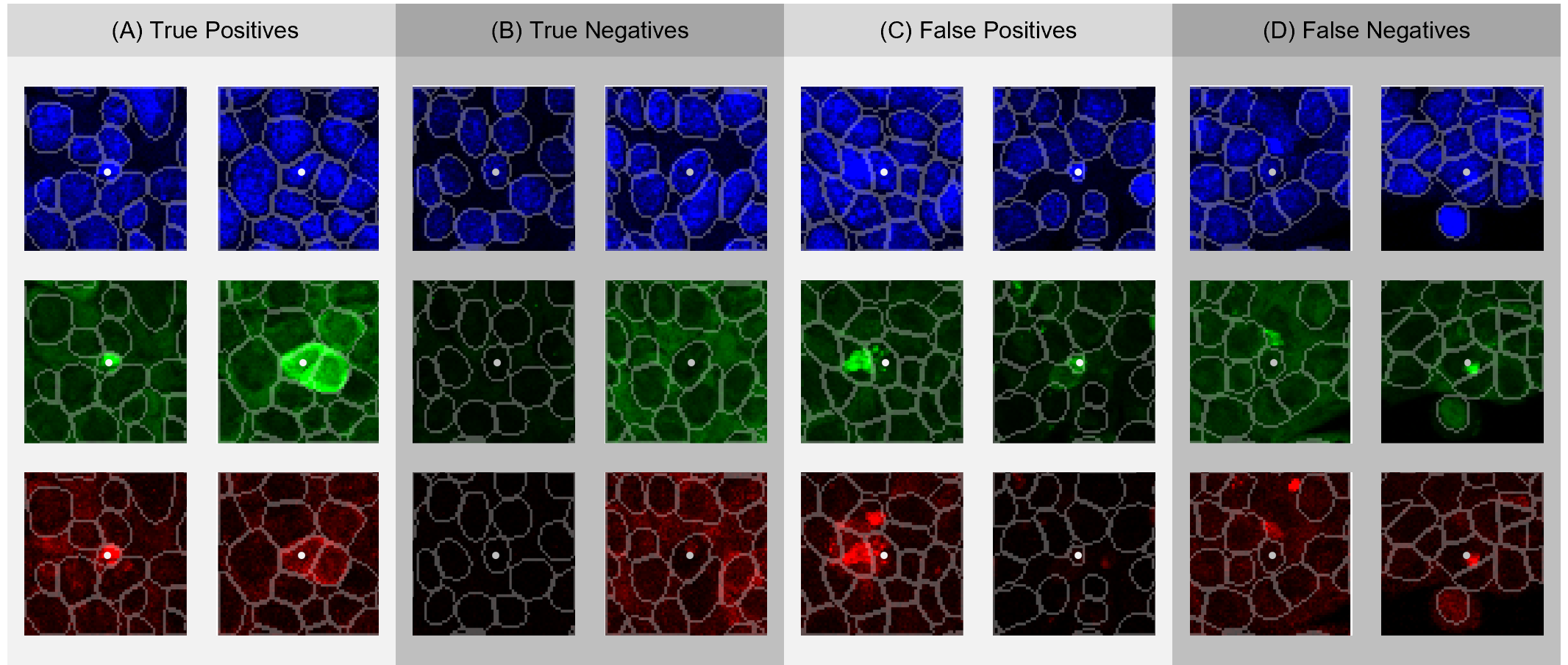}
    \caption{Two examples each of TP, TN, FP, and FN detections by the best performing \textbf{classification model} (Resnet-18). The cells of interest are centered in the tile and indicated by white dots. Unlike the heuristic visualization in Figure \ref{fig:heuristic_results}, where predictions were made on the entire image, the deep learning models classified isolated image patches; therefore, only the centrally located cell (white dot) represents the classification target, while surrounding cells (not classified) provide spatial context. The three rows correspond to the blue, green and red channels. The white outlines correspond to the Cellpose masks.}
    \label{fig:dl_results_resnet}
\end{figure*}


\section{Detection Methodology}

\subsection{Progressive Heuristics}
The heuristic methods were developed in successive stages, beginning with simple intensity thresholding, image-specific adaptation and progressively incorporating biological knowledge about fluorescence colocalization and artifact rejection. This progression reflects increasing model complexity while maintaining interpretable decision rules and \textbf{requiring no training}.

\subsubsection{Intensity Thresholding}

We first investigated whether dead cells could be identified using fluorescence intensity alone. Three method variants operated on per-cell mean intensities measured within Cellpose masks. 

The first used a fixed \textbf{percentile} threshold, classifying cells in the top 1\% (99th percentile) of both the green and red intensity distributions as dead. This represents the simplest assumption that apoptotic cells are consistently among the brightest cells in each image. 

The second estimated a threshold directly assuming a \textbf{bimodal} intensity distribution. A Gaussian kernel density estimate (KDE), with 512 grid points spanning the min to max intensity values, was computed from the per-cell intensities, and the first local minimum in the estimated density was used as the decision boundary. Local minima in the density curve were identified using relative minimum detection with an order parameter of 5 (requiring a dip between at least 5 data points). If no local minimum was detected, Otsu's threshold was used instead. Cells above the identified thresholds in both the green and red channels were counted as dead. 

The third approach identified dead cells based on statistical deviations from the population mean in fluorescence intensity. For each image, we calculate \textbf{z-scores} as $(intensity - mean) / standard\:deviation$ for both green and red channels, then classify cells with $z > 2.0$ in both channels as dead.

\subsubsection{Adaptive Gap Split}
The previous method relied on fixing thresholds and implicitly assumed that every image contained dead cells. In practice, many images contained none. We therefore developed an adaptive method that identified the largest separation in sorted per-cell mean intensities as the natural boundary between live and dead populations while also allowing the absence of any boundary at all. The algorithm computed \textit{gaps} between consecutive sorted intensity values, selected the maximum gap, and set its \textit{position} at the midpoint between the two adjacent intensities. If this gap was less than 10\% of the image's intensity range (either green or red channel), the image was deemed to have no dead cells on account of an absence of bimodality. Similarly, the gap cannot sit down in the background noise. If the gap threshold did not sit in the 75th percentile (either green or red channel), the image was again assumed to have no dead cells. The rest of the images were deemed to contain dead cells, identified with intensities above the gap's position.

\subsubsection{Colocalization-Aware Preprocessing}

Intensity alone is insufficient because apoptotic fluorescence must spatially follow the nuclear morphology. We therefore investigated two method variants that incorporated information from the blue nuclear channel before thresholding. 

The first calculated a blue-weighted intensity where a \textbf{normalized nuclear/cellular intensity} was multiplied pixelwise with the green and red channels before computing per-cell mean intensities, 
thereby emphasizing fluorescence originating from the nucleus. Cells exceeding the 99th percentile of the resulting scores in each of the green and red channel were classified as dead.

The second replaced simple blue-weighting approach with computed \textbf{Pearson correlations} between blue and green/red pixel intensities within each Cellpose mask. For each mask, we computed Pearson correlations between blue (nucleus) and green/red pixel values: $r_{g,b} = \mathrm{corr}(G, B)$ and $r_{r,b} = \mathrm{corr}(R, B)$. Cells with positive correlations ($r > 0$) were marked as dead if their mean green and mean red intensities exceeded their 99th percentiles among the positively correlated cells.

\subsubsection{Domain Expert-Informed Pipeline}
Although adaptive metrics reduce dependence on fixed thresholds, and colocalization ensures we only use relevant stain signals, the pipeline remained susceptible to noisy artifacts near cellular boundaries or bright/blotchy spots from debris or noise. Guided by domain knowledge, we built upon colocalization-aware preprocessing method through an iterative refinement process.

First, we tried incorporating and seeing the affect of adding a \textbf{coefficient of variation} (CV) filter in the colocalization-aware preprocessing (Pearson) method: cells with $\mathrm{CV} = \sigma / \mu > 1.0$ were excluded from the dead-cell prediction set, effectively removing cells with highly concentrated intensity distributions while retaining uniformly bright apoptotic nuclei. Similarly, we checked the effect of applying \textbf{boundary erosion} (3 pixels)  to the segmentation masks before applying Pearson Correlation and CV filters, to hopefully reduce false positives arising from edge artifacts, ensuring that only robust intracellular signal contributed to intensity measurements.

Finally, in order to be faithful to the domain scientist's point-of-view where the red channel was used solely as a secondary confirmation marker, we implemented a sophisticated prediction scheme. Cells exceeding the 99th percentile in green intensity (after erosion, Pearson, and CV filtering) were classified as dead. An additional set of cells in the 98$-$99 percentile range in green were classified as dead only if they also passed the red channel's 99th percentile threshold (calculated after erosion, Pearson, and CV filtering on the red channel).


\subsection{Classification}
Detecting rare events such as a single dead cell among hundreds of healthy ones represents an extreme class imbalance classification problem. We employed both traditional and deep learning based classification approaches.

\subsubsection{Using Numerical Features}
We evaluated four traditional classifiers: \textbf{Logistic Regression} with balanced class weights and a maximum of 2000 iterations, \textbf{Balanced Random Forest} with 500 estimators using \textit{balanced\_subsample} weighting, \textbf{XGBoost} with a \textit{logloss} objective and 500 estimators (no label encoder), and \textbf{LightGBM} with 500 estimators. Models were trained on a total of 27 features extracted from the three-channel cell masks and image patches as detailed in Table \ref{tab:feature table}. These consisted of per-channel statistical features - mean, median, maximum, and minimum intensity for each channel, and a Coefficient of Variation value for the cell mask. 

The 27 numerical features detailed in Table \ref{tab:feature table} were selected following preliminary feature-reduction trials. The candidate pool comprised over 60 features, including intermediate quartiles, standard deviations, and inter-quartile ranges computed across the cell mask, image patch, and global image scopes. Spatial features such as cell shape and area, were intentionally excluded from this feature set under the premise that complex geometric attributes are more robustly modeled by deep learning architectures. Furthermore, domain experts advised that morphological dimensions do not correlate with apoptotic indicators in this target distribution.

All features were standardized using \textit{StandardScaler} prior to training. Decision thresholds were selected using group-aware five-fold cross-validation to maximize the macro-F1 score. Training used 80/20 train/test splits (the test set contained 53 dead cells) on the dataset shown in Table \ref{tab:data_characteristics} (h-k).

\begin{table}[b]
\centering
\scriptsize
\caption{Scope and Count of Numerical Features}
\label{tab:feature table}
\begin{tabular}{llr}
\toprule
\textbf{Scope} & \textbf{Intensity Feature} & \textbf{Counts} \\
\midrule
\multirow{3}{*}{Cell} & mean  & 3 \\
& median  & 3 \\
& maximum  & 3 \\
& minimum  & 3 \\
& coefficient of variation & 3 \\
\midrule
\multirow{3}{*}{Image Patch} & mean  & 3 \\
& median  & 3 \\
& maximum  & 3 \\
& minimum  & 3 \\
\midrule
Total Features & & 27 \\
\bottomrule
\end{tabular}
\end{table}

\subsubsection{Using Image Features}
We fine-tuned \textbf{ResNet-18} and \textbf{EfficientNet-B0} architectures pretrained on ImageNet on the square image patches (all four variants) centered at segmented cell centroids (see Section \ref{sec:data_prep_patches} for details). Pixel values were first scaled to [0, 1] by dividing by 65535 as mentioned before, then normalized using ImageNet statistics ($\mu = [0.485, 0.456, 0.406]$, $\sigma = [0.229, 0.224, 0.225]$) which is important for pretrained models \cite{deng2009imagenet}. Models used weighted binary cross-entropy loss with positive class weight computed as the inverse of dead-cell prevalence in the training set, AdamW optimization with learning rate $3 \times 10^{-4}$ and weight decay $10^{-4}$, mixed-precision training on GPU with channel-last memory format when available \cite{loshchilov2017decoupled}. Decision thresholds were selected via group-aware five-fold cross-validation on the training set to maximize macro-F1 score. Training used 80/20 train/test splits (the test set contained 53 dead cells) on the dataset shown in Table \ref{tab:data_characteristics} (h-k). Training was done purely on image patches: \textbf{No mask information or numerical features were used}.


\subsection{Anomaly Detection}

As a complementary direction, we considered anomaly detection models \textbf{trained only on live cells}. The underlying assumption was that dead cells form a minority and deviate from the appearance of healthy cells in either numerical feature or image space. In this setting, high reconstruction error or outlier scores are interpreted as evidence of abnormality. We used both numerical and image features for several anomaly detection models trained exclusively on live cells, using identical datasets and splits that were used with the classification models.

\subsubsection{Using Numerical Features}
We evaluated (\textbf{Isolation Forest} with 500 estimators and contamination matching training dead-cell prevalence (approximately 0.37\%), \textbf{PCA reconstruction} with explained variance of 95\%, \textbf{One-Class SVM} with RBF kernel and $\nu$ (the upper bound on the fraction of training errors) set to the training dead-cell fraction (0.0037). All features were standardized using \textit{StandardScaler}.

\subsubsection{Using Image Features}
We trained a \textbf{Convolutional Autoencoder} with three convolutional encoder layers and stride-2 convolutions followed by transposed decoder and a \textbf{Deep SVDD} model with 64-dimensional latent representation. Both models were trained for 20 epochs using AdamW optimization (learning rate $10^{-3}$, weight decay $10^{-5}$) with mixed-precision training on GPU when available. Anomaly scores corresponded to reconstruction error (squared difference between input and output) for the autoencoder, and squared Euclidean distance from the learned center in latent space for Deep SVDD. Thresholds were set at the 99.63 percentile of live cell anomaly scores (corresponding to the 0.37\% dead-cell prevalence). Training was done purely on image patches: \textbf{No mask information or numerical features were used}.

\begin{table*}[!ht]
\centering
\scriptsize
\setlength{\tabcolsep}{4pt}
\caption{Comparative Performance of the Four Computational Approaches. The best performing metric is shown in bold. The best performing metric in each family is underlined.}
\label{tab:results_table}
\begin{tabular}{p{0.15\linewidth} llccccc} 
\toprule[1.25pt]
\textbf{Family} & \textbf{Method} & \textbf{Variant} & \textbf{PR-AUC} & \textbf{Precision} & \textbf{Recall} & \textbf{F1-Score} & \textbf{Macro-F1} \\

\midrule[1.0pt]
\multirow{10}{=}{Progressive Heuristics}
& \multirow{3}{*}{Intensity Thresholding} & percentile (99\%) & n/a & 0.354 & 0.572 & \underline{0.437} & \underline{0.717} \\
& & distribution-based (bimodal) & n/a & 0.111 & 0.455 & 0.179 & 0.586 \\
& & distribution-based (z-score) & n/a & 0.156 & \underline{0.782} & 0.260 & 0.626 \\
\cmidrule[0.5pt](lr){2-8}
& Adaptive Gap Split & size 10\%, position 75\% & n/a & \underline{0.482} & 0.383 & 0.427 & 0.712 \\
\cmidrule[0.5pt](lr){2-8}
& \multirow{2}{*}{Colocalization-Aware Preprocessing} & nuclear normalized & n/a & 0.317 & 0.601 & 0.415 & 0.706 \\
& & Pearson Correlation & n/a & 0.478 & 0.222 & 0.303 & 0.651 \\
\cmidrule[0.5pt](lr){2-8}
& \multirow{3}{*}{Domain Expert-Informed Pipeline} 
& add Coeff. of Variation filter & n/a & 0.352 & 0.235 & 0.281 & 0.640 \\
& & add cell boundary erosion & n/a & 0.360 & 0.263 & 0.304 & 0.651 \\
& & use red as secondary marker & n/a & 0.210 & 0.432 & 0.282 & 0.639 \\

\midrule[1.0pt]
\multirow{16}{=}{Classification\hspace{0.3\linewidth} (using numerical features)}
& \multirow{4}{*}{Logistic Regression} & 32px unpadded & 0.490 & 0.545 & 0.340 & 0.419 & 0.708 \\
& ~ & 32px padded & 0.492 & 0.515 & 0.321 & 0.395 & 0.697 \\
& ~ & 64px unpadded & 0.524 & 0.489 & 0.434 & 0.460 & 0.729 \\
& ~ & 64px padded & 0.491 & 0.457 & 0.396 & 0.424 & 0.711 \\
\cmidrule[0.5pt](lr){2-8}
& \multirow{4}{*}{Balanced Random Forest} & 32px unpadded & 0.531 & 0.545 & 0.453 & 0.495 & 0.746 \\
& ~ & 32px padded & 0.521 & 0.545 & 0.453 & 0.495 & 0.746 \\
& ~ & 64px unpadded & 0.578 & 0.478 & \underline{0.604} & 0.533 & 0.766 \\
& ~ & 64px padded & 0.579 & 0.485 & \underline{0.604} & 0.538 & 0.768 \\
\cmidrule[0.5pt](lr){2-8}
& \multirow{4}{*}{XGBoost} & 32px unpadded & 0.606 & \underline{0.667} & 0.528 & 0.589 & 0.794 \\
& ~ & 32px padded & 0.581 & 0.553 & 0.491 & 0.520 & 0.759 \\
& ~ & 64px unpadded & \underline{0.614} & 0.592 & 0.547 & 0.569 & 0.783 \\
& ~ & 64px padded & 0.560 & 0.633 & 0.585 & \underline{0.608} & \underline{0.803} \\
\cmidrule[0.5pt](lr){2-8}
& \multirow{4}{*}{LightGBM} & 32px unpadded & 0.137 & 0.327 & 0.340 & 0.333 & 0.665 \\
& ~ & 32px padded & 0.546 & 0.565 & 0.491 & 0.525 & 0.762 \\
& ~ & 64px unpadded & 0.488 & 0.583 & 0.396 & 0.472 & 0.735 \\
& ~ & 64px padded & 0.506 & 0.571 & 0.453 & 0.505 & 0.752 \\

\midrule[1.0pt]
\multirow{8}{=}{Classification\hspace{0.3\linewidth} (using image features)}
& \multirow{4}{*}{ResNet-18} & 32px\ unpadded & 0.537 & 0.542 & 0.491 & 0.515 & 0.756 \\
&  & 32px\ padded & 0.542 & 0.538 & 0.528 & 0.533 & 0.766 \\
&  & 64px\ unpadded & 0.616 & 0.611 & 0.623 & \underline{\textbf{0.617}} & \underline{\textbf{0.808}} \\
&  & 64px\ padded & 0.562 & 0.366 & 0.774 & 0.497 & 0.747 \\
\cmidrule[0.5pt](lr){2-8}
& \multirow{4}{*}{EfficientNet-B0} & 32px\ unpadded & 0.272 & 0.000 & 0.000 & 0.000 & 0.499 \\
&  & 32px\ padded & 0.505 & 0.245 & \underline{\textbf{0.906}} & 0.386 & 0.690 \\
&  & 64px\ unpadded & \underline{\textbf{0.627}} & 0.609 & 0.528 & 0.566 & 0.782 \\
&  & 64px\ padded & 0.619 & \underline{\textbf{0.720}} & 0.340 & 0.462 & 0.730 \\

\midrule[1.0pt]
\multirow{12}{=}{Anomaly Detection\hspace{0.2\linewidth} (using numerical features)}
& \multirow{4}{*}{Isolation Forest} & 32px\ unpadded 
& 0.259 & 0.278 & 0.377 & 0.320 & 0.658 \\
& ~ & 32px\ padded 
& 0.244 & 0.254 & 0.321 & 0.283 & 0.640 \\
& ~ & 64px\ unpadded 
& 0.213 & 0.257 & 0.358 & 0.299 & 0.648 \\
& ~ & 64px\ padded 
& 0.214 & 0.233 & 0.321 & 0.270 & 0.633 \\
\cmidrule[0.5pt](lr){2-8}
& \multirow{4}{*}{One-class SVM} & 32px\ unpadded 
& 0.406 & 0.244 & \underline{0.585} & 0.344 & 0.670 \\
& ~ & 32px\ padded 
& \underline{0.460} & 0.237 & \underline{0.585} & 0.337 & 0.666 \\
& ~ & 64px\ unpadded 
& 0.396 & 0.186 & 0.509 & 0.273 & 0.634 \\
& ~ & 64px\ padded 
& 0.391 & 0.190 & 0.509 & 0.277 & 0.636 \\
\cmidrule[0.5pt](lr){2-8}
& \multirow{4}{*}{PCA Reconstruction} & 32px\ unpadded 
& 0.355 & 0.280 & 0.434 & 0.341 & 0.669 \\
& ~ & 32px\ padded 
& 0.328 & 0.277 & 0.434 & 0.338 & 0.667 \\
& ~ & 64px\ unpadded 
& 0.326 & \underline{0.300} & 0.453 & \underline{0.361} & \underline{0.679} \\
& ~ & 64px\ padded 
& 0.301 & 0.288 & 0.434 & 0.346 & 0.671 \\
\midrule[1.0pt]
\multirow{8}{=}{Anomaly Detection\hspace{0.2\linewidth} (using image features)}
& \multirow{4}{*}{Convolutional Autoencoder} & 32px\ unpadded 
& \underline{0.088} & 0.130 & 0.132 & 0.131 & 0.564 \\
& ~ & 32px\ padded 
& 0.085 & \underline{0.160} & \underline{0.151} & \underline{0.155} & \underline{0.576} \\
& ~ & 64px\ unpadded 
& 0.035 & 0.044 & 0.038 & 0.041 & 0.519 \\
& ~ & 64px\ padded 
& 0.034 & 0.067 & 0.057 & 0.061 & 0.529 \\
\cmidrule[0.5pt](lr){2-8}
& \multirow{4}{*}{Deep SVDD} & 32px\ unpadded 
& 0.006 & 0.000 & 0.000 & 0.000 & 0.498 \\
& ~ & 32px\ padded 
& 0.007 & 0.017 & 0.019 & 0.018 & 0.507 \\
& ~ & 64px\ unpadded 
& 0.006 & 0.000 & 0.000 & 0.000 & 0.498 \\
& ~ & 64px\ padded 
& 0.004 & 0.000 & 0.000 & 0.000 & 0.498 \\
\bottomrule[1.25pt]
\end{tabular}
\end{table*}


\section{Results}

The three computational families exhibited a clear performance hierarchy (Table \ref{tab:results_table}). Deep learning and traditional classification using numerical features achieved the strongest overall performance. ResNet-18 trained on 64-pixel unpadded patches obtained the highest dead-cell F1-score (0.617) and macro-F1 (0.808). Traditional classification methods were highly competitive, with XGBoost reaching an F1-score of 0.608. Heuristic methods showed moderate success, while anomaly detection models consistently underperformed the supervised methods. These findings highlight the trade-offs between representation capacity, reliance on explicitly labeled examples, and domain interpretability.

\subsection{Heuristics}

Among the heuristic methods, Intensity Thresholding using the 99th percentile achieved the highest performance, reaching an F1-score of 0.437 and macro-F1 of 0.717. Distribution-based thresholding methods performed worse, indicating that the fluorescence distributions frequently lack a well-defined separating minimum. This was further made clear by the next method that was tried: The Adaptive Gap Split relaxed the assumption that every image contains dead cells, achieving a decent F1 of 0.427, but it could neither surpass the simple fixed percentile thresholding nor was successful at correctly identifying images that contained zero dead cells. For the reported gap size and position settings, the Adaptive Gap Split could only correctly identify 73 out of 125 images that contained no dead cells. These results strongly indicate that the dataset lacks true bimodality. Figure \ref{fig:heuristic_results} shows example images to help the reader visualize what the image channels, ground truth annotations, segmentation masks, and model predictions look like for true and false positives and negatives.

Introducing colocalization information through Pearson correlation or nuclear-normalized intensity did not yield noticeable improvements or degradation in the performance, suggesting that simple measures of spatial agreement remain insufficient to fully separate genuine apoptotic staining from bright imaging artifacts. Similarly, incorporating a coefficient of variation filter and cell boundary erosion further could not improve the performance in this dataset, illustrating the severe trade-off between specificity and sensitivity when manually designing decision rules. However, given that the Domain Expert-Informed Pipeline did not significantly degrade the performance and that all heuristic parameters can be fine-tuned for a given dataset, we believe that this method should still be prescribed in semi-automated or human-in-the-loop systems, offering a transparent algorithm that machine-learning methods cannot match.

\subsection{Classification}

Among the classification models using numerical features, XGBoost achieved the highest performance with an F1-score of 0.608 using 64-pixel padded patches. Balanced Random Forest also demonstrated strong predictive capability, reaching an F1 of 0.538 under the same patch configuration. Increasing the patch size from 32 to 64 pixels somewhat improved the performance.  

Classification using image features outperformed all other computational families. ResNet-18 trained on 64-pixel unpadded patches achieved the highest overall performance (F1 = 0.617, macro-F1 = 0.808), while EfficientNet-B0 also demonstrated top results in the remaining three metrics (PR-AUC, Precision, and Recall). 64-pixel crops performed much more strongly, indicating that surrounding spatial context, when learned directly from raw image data, contributes highly useful discriminative information for identifying rare apoptotic events. Figure \ref{fig:dl_results_resnet} shows example images to help the reader visualize what the image channels, segmentation masks, and model predictions look like for true and false positives and negatives.

\subsection{Anomaly Detection}

Anomaly detection methods underperformed classification approaches. Among numerical feature-based models, PCA Reconstruction produced the strongest results with an F1 of 0.361 using 64-pixel unpadded patches. Image-based approaches, including convolutional autoencoders and Deep SVDD, failed to identify dead cells reliably or at all, yielding several zero or near-zero F1-scores. These findings suggest that apoptotic nuclei do not simply represent statistical outliers in either feature space or image space. Instead, they occupy a complex region that overlaps substantially with healthy populations, making anomaly detection highly difficult without explicit positive examples.

\section{Discussion}

We systematically compared a spectrum of computational strategies, from interpretable heuristics to traditional and deep-learning based classification and anomaly detection, for a very challenging fluorescence microscopy classification problem. Note that heuristic models need no training, numerical based models used no images (spatial information) in training, image based models used no mask or numerical features in training, and anomaly detection models were trained exclusively on numerical/image data belonging to live cells.

\subsection*{Why is automated dead-cell detection difficult?}

Automated dead-cell detection represents a particularly challenging computer vision problem because multiple independent sources of uncertainty accumulate throughout the analysis pipeline. First, the quality of downstream classification is limited by segmentation accuracy. Missed or merged nuclei or imperfect boundaries directly affect every subsequent measurement (Figure \ref{fig:heuristic_results} b, d). Second, fluorescence intensity alone is an unreliable indicator: In practice and in our data, fluorescence images contain non-nuclear background noise for the apoptotic marker as well (Figure \ref{fig:heuristic_results}a). Because the fluorescent dyes strictly target the nuclei rather than the whole cell, apoptotic nuclei appear uniformly bright when dead; however, overall staining intensity varies considerably between images, and many acquisitions contain no dead cells at all. Bright fluorescence can also arise from debris or nonspecific staining, producing signals that resemble apoptotic events (Figure \ref{fig:heuristic_results} a, c). Finally, the dataset presents an extreme class imbalance, with dead cells representing only approximately 0.37\% of all segmented targets. Such imbalance severely limits the statistical power available for learning while making false positives highly costly.

\subsection*{Heuristics versus supervised classification}

The progression observed across the heuristic methods provides an interesting perspective. Each successive heuristic attempted to encode additional biological knowledge by improving signal representation or rejecting artifacts. Although these refinements increased interpretability, they highlighted the difficulty of manually specifying all relevant decision rules. Supervised classification, particularly deep learning (ResNet-18, EfficientNet-B0) and gradient boosting (XGBoost), effectively learned these complex relationships directly from the data. Rather than relying on explicit intensity thresholds, these models captured the spatial information in the images and gracefully modeled the uniformly bright nature of the target nuclei alongside the contextual information surrounding them.

\subsection*{Practical implications}

Although fully automated detection of extremely rare events remains challenging, the presented pipeline is immediately applicable as a decision-support tool. Rather than replacing manual inspection, the best-performing models can prioritize a small subset of candidate cells for expert review, substantially reducing annotation effort during large-scale screening. This human-in-the-loop workflow accelerates dataset annotation, improves consistency, and facilitates the creation of larger labeled datasets.


\section{Conclusions}

This study presented a comprehensive evaluation of heuristic algorithms, traditional machine learning, deep learning, and anomaly detection methods for automated dead-cell detection in fluorescence microscopy. Using a highly imbalanced dataset containing over 65,000 segmented targets and only 243 annotated dead cells, we systematically compared the efficacy of each computational strategy. Classification using numerical features (traditional) and image features (deep-learning) achieved the highest performance, with ResNet-18 (F1 = 0.617) and XGBoost (F1 = 0.608) significantly outperforming manually designed heuristics and anomaly detection methods. While deterministic rules and isolation algorithms struggle to separate the uniformly bright apoptotic nuclei from overlapping fluorescence artifacts, supervised models successfully leverage surrounding spatial context to improve classification. Ultimately, while fully autonomous detection remains challenging due to inherent staining variability and segmentation limits, these top-performing models provide an immediately viable human-in-the-loop solution. Automatically prioritizing candidate nuclei for expert review can drastically reduce manual effort, ensuring highly scalable and consistent analysis in biomedical imaging applications.


\section*{Code availability}
\label{sec:data_code_availability}

The complete implementation of various computational approaches is publicly available at \url{https://github.com/pv-is-nrt/pub-dead-cell-detection}.


\section*{Acknowledgment}

This work was supported in part by the National Science Foundation under awards 1946391, the National Institute of General Medical Sciences (NIGMS) of National Institutes of Health (NIH) under awards P20GM139768, and the Arkansas Integrative Metabolic Research Center (AIMRC) at University of Arkansas.


\bibliographystyle{IEEEtran}
\bibliography{IEEEabrv, references}




\end{document}